\documentclass[10pt,twocolumn,letterpaper]{article}

\usepackage{cvpr}              

\usepackage{subcaption}
\usepackage{pifont}
\usepackage{makecell}
\usepackage{booktabs}
\usepackage[table]{xcolor}
\usepackage{graphicx}
\newcommand{\cmark}{\ding{51}} 
\newcommand{\xmark}{\ding{55}} 
\newcommand{\best}[1]{\textcolor{red}{#1}}
\newcommand{\secondbest}[1]{\textcolor{blue}{#1}}

\usepackage{multirow}
\usepackage{array}
\usepackage[table]{xcolor}  
\usepackage{pifont}
\usepackage{booktabs}  
\usepackage{diagbox}  

\definecolor{cvprblue}{rgb}{0.21,0.49,0.74}
\usepackage[pagebackref,breaklinks,colorlinks,allcolors=cvprblue]{hyperref}

\def\paperID{}
\def\confName{CVPR}
\def\confYear{2026}

\title{VisualAD: Language-Free Zero-Shot Anomaly Detection via Vision Transformer}

\author{
\parbox{\textwidth}{\centering
Yanning Hou$^{3,2}$ \quad
Peiyuan Li$^{2}$ \quad
Zirui Liu$^{2}$ \quad
Yitong Wang$^{2}$ \\
Yanran Ruan$^{2}$ \quad
Jianfeng Qiu$^{2}$ \quad
Ke Xu$^{1,2*}$ \\[0.6em]
$^{1}$ State Key Laboratory of Opto-Electronic Information Acquisition and Protection Technology, Anhui University, Hefei, China \\
$^{2}$ School of Artificial Intelligence, Anhui University, Hefei, China \\
$^{3}$ College of Intelligence Science and Technology, National University of Defense Technology, Changsha, China
}
}
\newcommand\blfootnote[1]{%
  \begingroup
  \renewcommand\thefootnote{}\footnote{#1}%
  \addtocounter{footnote}{-1}%
  \endgroup
}

\begin{document}

\maketitle
\begin{abstract}
Zero-shot anomaly detection (ZSAD) requires detecting and localizing anomalies without access to target-class anomaly samples. Mainstream methods rely on vision–language models (VLMs) such as CLIP: they build hand-crafted or learned prompt sets for normal and abnormal semantics, then compute image–text similarities for open-set discrimination. While effective, this paradigm depends on a text encoder and cross-modal alignment, which can lead to training instability and parameter redundancy. This work revisits the necessity of the text branch in ZSAD and presents VisualAD, a purely visual framework built on Vision Transformers. We introduce two learnable tokens within a frozen backbone to directly encode normality and abnormality. Through multi-layer self-attention, these tokens interact with patch tokens, gradually acquiring high-level notions of normality and anomaly while guiding patches to highlight anomaly-related cues. Additionally, we incorporate a Spatial-Aware Cross-Attention (SCA) module and a lightweight Self-Alignment Function (SAF): SCA injects fine-grained spatial information into the tokens, and SAF recalibrates patch features before anomaly scoring. VisualAD achieves state-of-the-art performance on 13 zero-shot anomaly detection benchmarks spanning industrial and medical domains, and adapts seamlessly to pretrained vision backbones such as the CLIP image encoder and DINOv2. Our code is publicly available at \url{https://github.com/7HHHHH/VisualAD}.
\end{abstract}

\blfootnote{* Corresponding author.}
\section{Introduction}
\label{sec:intro}\begin{figure}[!htb]
    \centering    
 \includegraphics[width=0.48\textwidth]{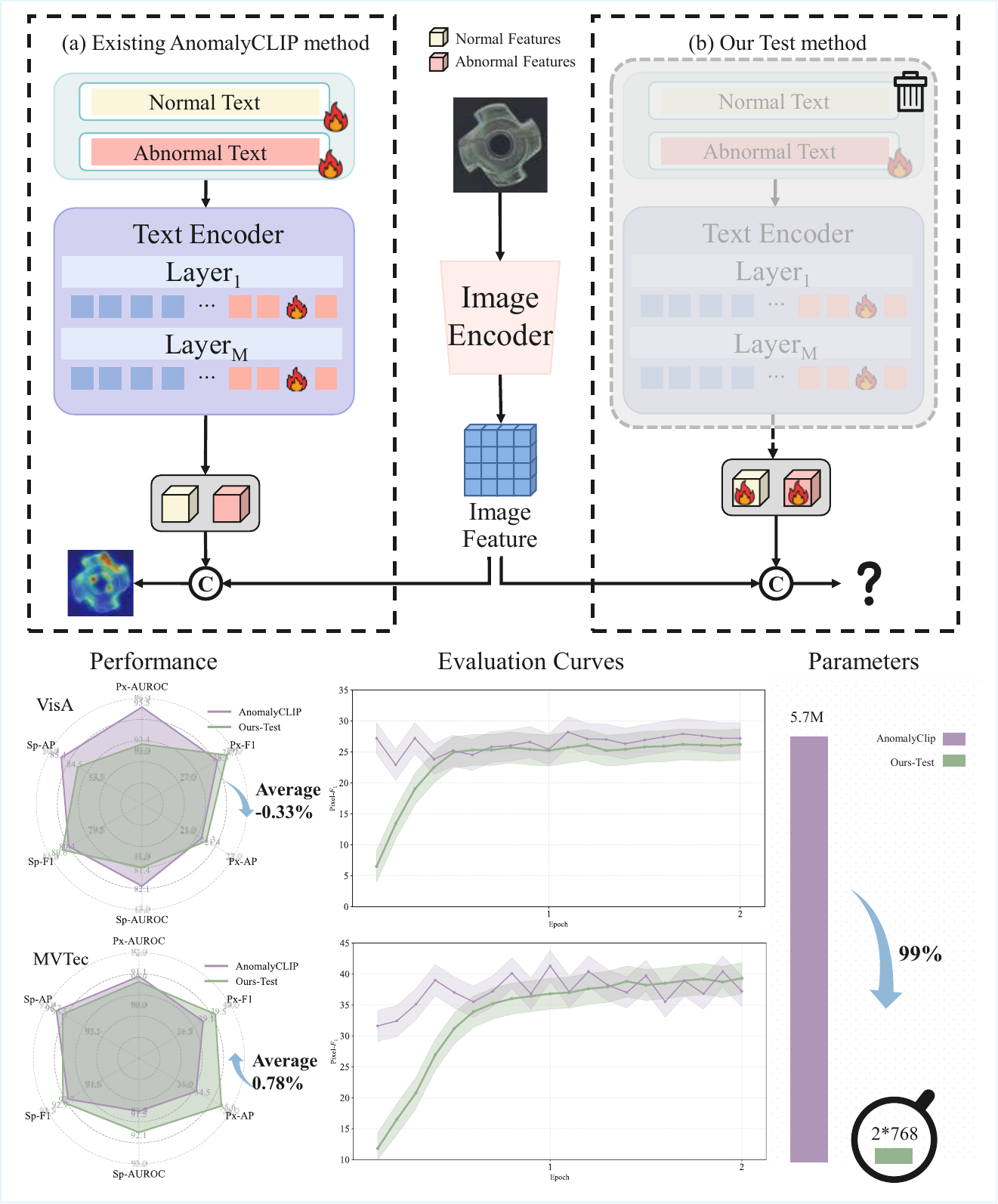} 
    \caption{Exploratory study motivating this work. (a) AnomalyCLIP derives normal/abnormal prototypes via trainable text prompts and a text encoder. (b) A purely visual variant removes the text branch and directly learns two visual prototypes, achieving comparable or slightly better results on VisA and MVTec with ~99\% fewer parameters. Radar and line plots show that our variant maintains similar accuracy but with much smoother evaluation curves, whereas AnomalyCLIP oscillates noticeably.}
    \label{fig:m1}
\end{figure} 

Anomaly detection~\cite{AD1,AD2,AD3,AD4} in images plays a central role in safety-critical domains such as industrial inspection and medical diagnosis. The objective is to identify deviations from expected patterns at either the image or pixel level, from subtle surface defects on manufactured parts to early pathological changes in medical scans. Most established methods follow unsupervised~\cite{unsupervised1,unsupervised2,unsupervised3,unsupervised4} or semi-supervised~\cite{PromptAD,semi} paradigms that assume access to category-specific normal images, labeled anomalies, or both during training. In realistic deployments, however, a cold-start situation is common: collecting sufficient, clean normal data for every new product line or disease category is costly and sometimes impractical. As a result, these methods may be difficult to scale to long-tail categories and are often brittle under domain shifts between carefully curated training data and noisy real-world environments. Zero-shot anomaly detection (ZSAD)~\cite{AdaCLIP,AnomalyCLIP,AdaptCLIP,MetaUASUA} seeks to alleviate these limitations by detecting anomalies in previously unseen categories without using any in-category images for training, thereby decoupling anomaly detection from the per-category data collection burden.

Against this backdrop, large-scale pretrained models~\cite{DINO,DINOv2} offer a practical route to ZSAD. Pretrained vision–language models (VLMs)~\cite{CLIP,BLIP-2,BLIP} such as CLIP~\cite{CLIP} exhibit strong transferability by aligning visual and textual semantics in a shared embedding space. Leveraging this property, a line of work introduces prompts denoting normality and abnormality into the text encoder and contrasts the resulting text embeddings with image features to perform open-set anomaly detection. Within this paradigm, textual prompts are typically divided into two categories. The hand-crafted~\cite{APRIL-GAN,CLIP-AD,StackCLIP} category, exemplified by WinCLIP~\cite{WinCLIP}, relies on fixed templates that combine category context words with state descriptors to express normal and abnormal states; although training-free and easy to apply, such designs are sensitive to wording choices and may struggle to cover diverse anomaly patterns. The learnable category~\cite{AnomalyCLIP,AdaCLIP,AdaptCLIP}, exemplified by AnomalyCLIP~\cite{AnomalyCLIP}, parameterizes contextual terms as trainable vectors and optimizes them on a small auxiliary dataset, yielding object-agnostic representations of normality and abnormality that can generalize across categories. Despite their differences, these methods suggest a simple principle: anomalies can be described by two sets of reference features corresponding to normality and abnormality.

However, this principle raises a question: if the final decision is governed by only two sets of latent vectors, normal and anomaly, is the language modality truly indispensable? From a purely visual standpoint, anomalies appear as structural or statistical deviations in texture, shape, or color, all of which can be captured within the visual domain. To probe this hypothesis, we conducted a small-scale study by modifying AnomalyCLIP~\cite{AnomalyCLIP}: we removed the text encoder and directly optimized two learnable vectors representing normality and abnormality. The resulting detection performance showed negligible degradation, while the number of trainable parameters dropped by more than 99\%. As illustrated in Fig.~\ref{fig:m1}, the evaluation curves across training epochs reveal a clear difference in generalization behavior: our visual-only variant improves steadily and maintains a smooth upward trend, whereas the original AnomalyCLIP displays pronounced fluctuations across epochs. This finding suggests that, in CLIP-based ZSAD pipelines, textual prompts might serve primarily as an indirect route for shaping a pair of discriminative visual prototypes, rather than providing essential semantic grounding.

Motivated by this observation, we propose VisualAD, a purely visual framework for zero-shot anomaly detection built on frozen Vision Transformers. Instead of relying on a text encoder, VisualAD introduces two learnable tokens, an anomaly token and a normal token, directly into the ViT token sequence. Through multi-layer self-attention, these tokens interact with patch tokens, gradually acquiring high-level notions of normality and anomaly while guiding patches to emphasize anomaly-related cues. To align the high-level semantic tokens with fine-grained image features across layers, we employ a Spatial-Aware Cross-Attention (SCA) module that provides explicit spatial cues and local detail to the tokens, together with a lightweight Self-Alignment Function (SAF), implemented as a small MLP, that recalibrates patch features before token–patch alignment.  Unlike methods that align directly to patch-level features, SCA dynamically updates the tokens using fine-grained visual evidence, which improves ZSAD performance. We then synthesize the anomaly map by integrating alignment outcomes between multiple updated tokens and recalibrated patch-level features across layers, and obtain the image-level anomaly decision from the top one percent of pixels with the highest anomaly scores.

In summary, our contributions are as follows:
\begin{itemize}
\item We revisit the necessity of text in zero-shot anomaly detection and show that discriminative anomaly features can be learned purely from visual cues.
\item We introduce VisualAD, a ViT-only framework for zero-shot anomaly detection that injects two learnable tokens into a frozen backbone, allowing them to interact with patch tokens through multi-layer self-attention to encode normality and abnormality.
\item We propose the SCA and SAF modules, where SCA injects explicit spatial evidence into the tokens and SAF recalibrates patch features, enabling stable multi-layer alignment and improved localization.
\item Extensive experiments across industrial and medical benchmarks demonstrate strong zero-shot generalization to unseen categories and datasets.
\end{itemize}

\section{Related work}
\label{sec:formatting}
\subsection{Unsupervised Anomaly Detection.}
Unsupervised anomaly detection seeks to identify anomalies using only normal training images. Existing methods fall into three categories: embedding-based, discriminative, and reconstruction-based approaches. Embedding-based methods, such as PaDiM~\cite{PaDiM}, PatchCore~\cite{PathCore}, and CSFlow~\cite{CS-Flow}, utilize features from pre-trained models to distinguish anomalies from normal samples. Discriminative methods, like CutPaste~\cite{CutPaste} and DRAEM~\cite{DRAEM}, introduce synthetic anomalies to convert the task into a supervised problem. Reconstruction-based~\cite{RE1,RE2,RE3} methods, including autoencoders~\cite{Auto1,Auto2,SoftRP} and GANs~\cite{GAN1,GAN2,GAN3}, assume anomalies yield higher reconstruction errors due to their absence in normal data. Recent knowledge distillation and feature reconstruction approaches balance efficiency and performance by aligning a student network with a pre-trained teacher. However, these methods struggle with unseen anomaly types, requiring extensive normal data collection and retraining for new scenarios, which limits their practical adaptability.

\subsection{Zero-shot Anomaly Detection.}
In industrial anomaly detection, pre-trained vision models~\cite{VIT,BLIP,CLIP,SAM,DINO} show strong generalization and feature extraction. Zero-shot approaches fall into two families. The first is training-free: WinCLIP~\cite{WinCLIP} aligns multi-granularity features but needs repeated encodings; SAA~\cite{SAA} couples Grounding DINO~\cite{GroundingDINO} with SAM~\cite{SAM} for precise localization and segmentation at higher computational cost. The second family keeps the target domain zero-shot while training lightweight modules on auxiliary data to align vision and text. AnomalyCLIP~\cite{AnomalyCLIP} uses object-agnostic prompts with glocal optimization but relies on a dual branch; AdaCLIP~\cite{AdaCLIP} and VCP-CLIP~\cite{VCP} inject visual cues into text prompts, improving adaptability with added complexity; CLIP-AD~\cite{CLIP-AD} and CLIP-SAM~\cite{CLIP-SAM} further tighten alignment or add segmentation priors. AA-CLIP~\cite{AA} adopts a two-stage scheme with anomaly and normal text anchors plus lightweight visual adapters; AdaptCLIP~\cite{AdaptCLIP} learns compact visual, textual, and prompt–query adapters for cross-domain transfer; Bayes-PFL~\cite{Bayes-PFL} models a distribution over prompts to capture diverse anomaly semantics. In contrast, VisualAD removes the text branch and learns high-level anomaly cues purely from vision, achieving strong zero-shot results.
\section{Approach}
\label{sec:approach}
\begin{figure}[t]
    \centering
    \includegraphics[width=0.92\linewidth]{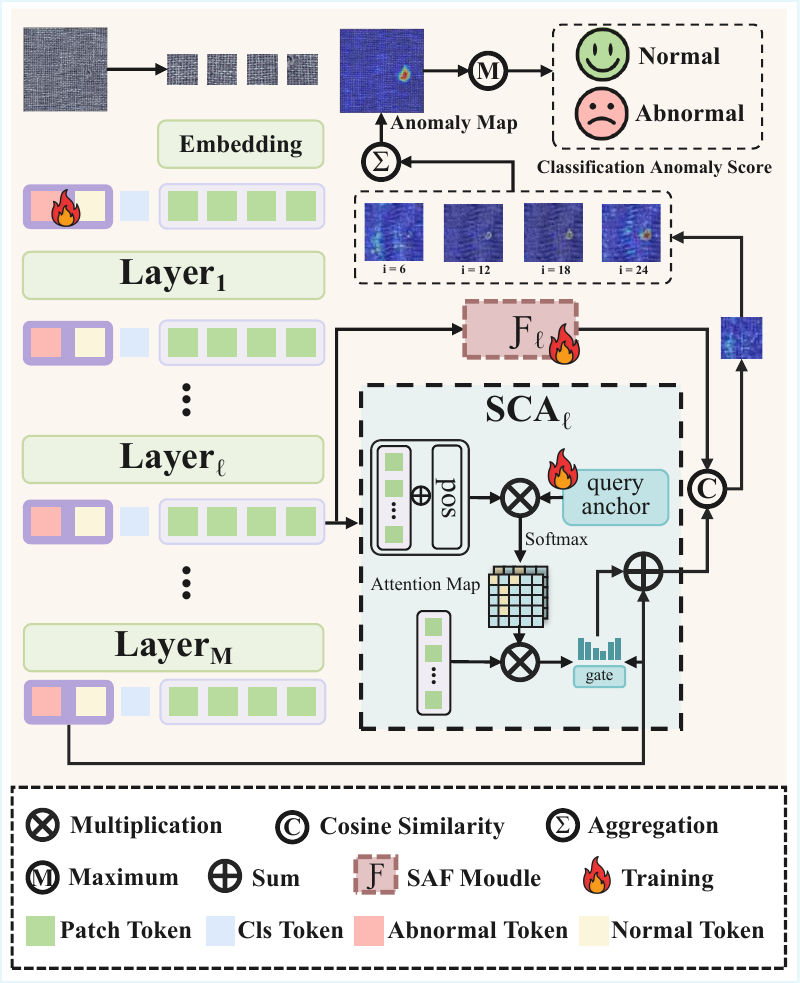}
    \caption{\textbf{Overview of VisualAD.} Two learnable global tokens (anomaly and normal) are inserted into a frozen ViT. For intermediate layers \( \ell\), SCA uses a few anchor queries with positional encoding and token-guided gating to aggregate localized spatial evidence, yielding enhanced tokens \( \tilde{\mathbf{t}}_a^{(\ell)}, \tilde{\mathbf{t}}_n^{(\ell)} \). In parallel, each layer's patch features are recalibrated by a SAF, denoted as \( \mathcal{F}_\ell \). The cosine-similarity difference between enhanced tokens and recalibrated patches forms layer-wise anomaly maps, which are upsampled and summed to obtain the final anomaly map; the image-level score is the mean of the top-\(k\) responses.}
    \label{fig:framework}
\end{figure}
\subsection{Problem Setting}
\label{sec:setting}
We follow the zero-shot anomaly detection (ZSAD) protocol. The model is trained with supervision on visible categories \(C_s\) drawn from industrial datasets and is directly evaluated on unseen categories \(C_u\) collected from other industrial or medical datasets. By definition \(C_u \cap C_s = \varnothing\); the auxiliary training and testing categories come from distinct datasets that exhibit substantial domain shifts. This setting evaluates cross-domain generalization without any category-specific fine-tuning.

As illustrated in Fig.~\ref{fig:framework}, we propose \textbf{VisualAD}, a zero-shot anomaly detection framework built upon frozen Vision Transformers. Given a backbone such as ViT-L/14@336px, we insert two learnable global tokens, the anomaly token \(\mathbf{t}_a\) and the normal token \(\mathbf{t}_n\), to directly encode abnormality and normality in the visual feature space. The input sequence is
\begin{equation}
\mathbf{z}_0 = [\mathbf{t}_a,\, \mathbf{t}_n,\, \mathbf{t}_c,\, \mathbf{p}_1,\, \ldots,\, \mathbf{p}_N],
\end{equation}
where \(\mathbf{t}_c\) is the class token, \(\{\mathbf{p}_i\}_{i=1}^N\) are image patch tokens, and all tokens are \(d\)-dimensional.

To capture diverse spatial and semantic cues, we extract features from a set of intermediate layers \(L=\{l_1,\ldots,l_K\}\) (default \(\{6,12,18,24\}\) for ViT-L/14). For each selected layer, we introduce a \textit{Spatial-Aware Cross-Attention} (SCA) module that aggregates localized spatial evidence from patch features and enhances the token representations. In parallel, a lightweight \textit{Self-Alignment Function} (SAF) recalibrates patch features at layer \(\ell\). The enhanced tokens and the SAF-recalibrated patches jointly produce layer-wise anomaly maps that are fused across layers. Pixel-level and image-level anomaly scores are obtained from the fused map. Training and inference share the same computation path with no text encoder or cross-modal alignment.


\subsection{Spatial-Aware Cross-Attention}
\label{sec:sca}

Global tokens encode high-level semantics but lack explicit spatial grounding. To inject localized evidence while keeping the ViT pipeline intact, we apply a single SCA module at each selected layer~$\ell$. Given patch features $\mathbf{P}_\ell \in \mathbb{R}^{B\times N\times d}$, where $B$ is the batch size, $N$ is the number of patches, and $d$ is the feature dimension, we first augment them with a layer-specific learnable positional encoding $\mathbf{E}_{\text{pos}}^{(\ell)}$:
\begin{equation}
    \mathbf{P}_\ell^{\text{pos}} = \mathbf{P}_\ell + \mathbf{E}_{\text{pos}}^{(\ell)}.
\end{equation}

allowing the module to distinguish spatially distinct regions.

To aggregate spatial cues efficiently, SCA employs $m$ learnable anchor queries $\mathbf{Q}_{\text{anchor}}\in\mathbb{R}^{m\times d}$ with $m\ll N$. For clarity, we omit the batch dimension in the following and denote $\mathbf{P}_\ell^{\text{pos}}\in \mathbb{R}^{N\times d}$. The cross-attention weights and anchor-aggregated features are computed as
\begin{equation}
\begin{aligned}
\mathbf{A}_\ell &= \mathrm{softmax}\!\left(
        \frac{\mathbf{Q}_{\text{anchor}}(\mathbf{P}_\ell^{\text{pos}})^\top}{\sqrt{d}}
    \right),
\qquad
\mathbf{U}_\ell = \mathbf{A}_\ell\, \mathbf{P}_\ell .
\end{aligned}
\end{equation}
where $\mathbf{A}_\ell \in \mathbb{R}^{m\times N}$ is the spatial attention distribution and $\mathbf{U}_\ell \in \mathbb{R}^{m\times d}$ contains the anchor-aggregated features, whose $i$-th row is denoted by $\mathbf{a}_i$.

Next, SCA adapts the anchor features to each token via a token-guided gating mechanism. We compute a gate vector
\begin{equation}
    \mathbf{g}(\mathbf{t}) = \sigma(W_g \mathbf{t}) \in \mathbb{R}^{m},
\end{equation}
where $\mathbf{t} \in \{\mathbf{t}_a,\,\mathbf{t}_n\}$ is a global token before enhancement, and $W_g$ is shared across anchors. The enhanced token is
\begin{equation}
    \tilde{\mathbf{t}}_\ell 
    = \mathbf{t} 
    + \alpha \sum_{i=1}^{m} g_i(\mathbf{t})\, \mathbf{a}_i ,
\end{equation}
with $\alpha$ a learnable residual scale and $g_i(\mathbf{t})$ the $i$-th entry of $\mathbf{g}(\mathbf{t})$. This formulation preserves the semantic role of global tokens while selectively injecting spatial information in an anchor-specific manner.

We instantiate SCA independently for each $\ell \in \mathcal{L}$, producing $\tilde{\mathbf{t}}_{a}^{(\ell)}$ and $\tilde{\mathbf{t}}_{n}^{(\ell)}$ that are subsequently matched with the recalibrated patch features to form the layer-wise anomaly maps. Since attention and gating are recomputed for every image, SCA dynamically adapts the abnormality sensitivity of the global tokens to the local structure of each test sample.

\subsection{Self-Alignment Function and Anomaly Scoring}
\label{sec:anomaly_score}

Beyond SCA, we refine patch representations using a learnable \emph{Self-Alignment Function (SAF)}.
For each selected layer~$\ell$, the SAF is implemented as a one-hidden-layer MLP with hidden
dimension equal to the input size:
\begin{equation}
    \hat{\mathbf{P}}_\ell = \mathcal{F}_\ell(\mathbf{P}_\ell),
\end{equation}
where $\mathcal{F}_\ell$ denotes the SAF at layer~$\ell$. This nonlinear recalibration aligns
patch features with the evolving normal and anomaly tokens.

Patch-level anomaly scores are then computed through a self-aligned cosine contrast. Using
$\ell_2$-normalized features ($\bar{\mathbf{x}}=\mathbf{x}/\|\mathbf{x}\|_2$), the score of
the $i$-th patch at layer~$\ell$ is
\begin{equation}
    s_i^{(\ell)} =
    \big\langle \bar{\hat{\mathbf{p}}}_i^{(\ell)},\, \bar{\mathbf{t}}_a^{(\ell)} \big\rangle
    -
    \big\langle \bar{\hat{\mathbf{p}}}_i^{(\ell)},\, \bar{\mathbf{t}}_n^{(\ell)} \big\rangle.
\end{equation}

Scores are reshaped into spatial maps $\mathbf{H}_\ell \in \mathbb{R}^{H\times W}$ and
upsampled to the input size. Multi-layer fusion yields the final anomaly map:
\begin{equation}
    \mathbf{H} = \sum_{\ell\in\mathcal{L}} \mathbf{H}_\ell.
\end{equation}

The image-level anomaly score~$S$ is computed by averaging the top-$k$ most anomalous pixels,
where $k = \lfloor 0.01\, HW \rceil$ denotes the floor of 1\% of all pixels:
\begin{equation}
    S = \frac{1}{k}\sum_{i=1}^{k} \mathbf{H}_{(i)},
\end{equation}
with $\mathbf{H}_{(i)}$ the $i$-th largest value in $\mathbf{H}$. This completes the construction
of both the anomaly map and the corresponding image-level score in VisualAD.

\subsection{Training Objective}
\label{sec:training_objective}
We jointly optimize image-level classification, pixel-level segmentation, and token contrastive separation under a unified objective, while keeping the ViT backbone frozen; only the anomaly/normal tokens \(\mathbf{t}_a,\mathbf{t}_n\), the SCA modules, and the per-layer transformations \(\{\mathcal{F}_\ell\}\) are updated. For image-level supervision, we apply binary cross-entropy to the score \(S\):
\begin{equation}
\mathcal{L}_{\mathrm{cls}}
= -\Big[y\log\sigma(S) + (1-y)\log\!\big(1-\sigma(S)\big)\Big].
\end{equation}
For pixel-level supervision, we combine Focal Loss and Dice Loss on each selected layer map, where \(\hat{\mathbf{M}}^{(\ell)}=\sigma(\mathbf{H}_\ell)\) denotes the predicted mask and \(\mathbf{M}\) is the binary ground truth:
\begin{equation}
\mathcal{L}_{\mathrm{seg}}
= \sum_{\ell\in\mathcal{L}}
\left(\mathcal{L}_{\mathrm{focal}}^{(\ell)} + \mathcal{L}_{\mathrm{dice}}^{(\ell)}\right).
\end{equation}

To explicitly separate the anomaly and normal tokens, we impose a cosine-margin penalty at the deepest layer index $L$ (the maximum element in $\mathcal{L}$): 
\begin{equation}
\mathcal{L}_{\mathrm{ctr}}
= \max\Big(0,\ \big\langle \bar{\mathbf{t}}_a^{(L)}, \bar{\mathbf{t}}_n^{(L)} \big\rangle + \tau\Big),\quad \tau=0.5,
\end{equation}
which encourages their similarity to fall below \(-0.5\), corresponding to an angular distance greater than \(120^\circ\). The overall loss is
\begin{equation}
\mathcal{L} = \mathcal{L}_{\mathrm{cls}} + \mathcal{L}_{\mathrm{seg}} + \mathcal{L}_{\mathrm{ctr}}.
\end{equation}
Training and inference strictly share the same pipeline, including the layer set \(\mathcal{L}\), SCA, the transformations \(\mathcal{F}_\ell\), cross-layer fusion, and top-\(k\) aggregation, with no text branch or cross-modal alignment.
\begin{table*}[htbp]
  \centering
  \caption{Comparison of zero-shot anomaly detection methods on industrial and medical benchmarks.
For each dataset, we report image-level metrics (AUROC, $F_1$-max, AP) and pixel-level metrics (AUROC, $F_1$-max, AP, PRO).
VisualAD(CLIP) and VisualAD(DINOv2) denote our method instantiated with CLIP and DINOv2 backbones, respectively.
The best and second-best results for each metric and dataset are highlighted in \best{red} and \secondbest{blue}.}
  \renewcommand{\arraystretch}{1.5}%
  \resizebox{2.1\columnwidth}{!}
{%
    \begin{tabular}{cccccccccc}
    \toprule
    \multirow{2}[4]{*}{Domain} & \multirow{2}[4]{*}{Metric} & \multirow{2}[4]{*}{Dataset} & WinCLIP~\cite{WinCLIP} & APRIL-GAN~\cite{APRIL-GAN} & CLIP-AD~\cite{CLIP-AD} & AnomalyCLIP~\cite{AnomalyCLIP} & AdaCLIP~\cite{AdaCLIP} & \cellcolor[rgb]{ .906,  .902,  .902}VisualAD(CLIP~\cite{CLIP}) & \cellcolor[rgb]{ .906,  .902,  .902}VisualAD(DINOv2~\cite{DINOv2}) \\
    \cmidrule{4-10}
          &       &       & ViT-L/14@336px & ViT-L/14@336px & ViT-L/14@336px & ViT-L/14@336px & ViT-L/14@336px & \cellcolor[rgb]{ .906,  .902,  .902}ViT-L/14@336px & \cellcolor[rgb]{ .906,  .902,  .902}ViT-L/14 \\
    \midrule
    \multirow{12}[4]{*}{Industrial}
          & \multicolumn{1}{c}{\multirow{6}[2]{*}{\makecell{Image-level\\(AUROC, $F_1$-max,\\ AP)}}}
          & MVTec-AD~\cite{MVTec}  & (90.4, \secondbest{92.7}, 95.6)  & (86.1, 90.4, 93.6)  & (74.1, 86.3, 88.1)  & (91.6, \secondbest{92.7}, 96.2)  & (\secondbest{92.0}, \secondbest{92.7}, \secondbest{96.4})  & \cellcolor[rgb]{ .906,  .902,  .902}(\best{92.2}, \best{93.2}, \best{96.7})  & \cellcolor[rgb]{ .906,  .902,  .902}(90.1, 92.4, 94.8) \\
          &        & VisA~\cite{visa}   & (75.6, 78.2, 78.8)  & (77.4, 78.6, 80.9)  & (66.2, 74.3, 71.4)  & (81.0, 80.3, 84.4)  & (79.7, 79.63, 83.2)  & \cellcolor[rgb]{ .906,  .902,  .902}(\best{84.7}, \best{82.5}, \best{87.6})  & \cellcolor[rgb]{ .906,  .902,  .902}(\secondbest{83.1}, \secondbest{81.4}, \secondbest{86.8}) \\
          &        & BTAD~\cite{BTAD}   & (68.2, 67.8, 70.9)  & (73.7, 68.7, 69.9)  & (66.7, 65.9, 67.3)  & (88.7, 86.0, 90.6)  & (\secondbest{90.0}, \secondbest{87.2}, \secondbest{91.5})  & \cellcolor[rgb]{ .906,  .902,  .902}(\best{94.9}, \best{93.9}, \best{97.0})  & \cellcolor[rgb]{ .906,  .902,  .902}(88.2, 84.7, 89.7) \\
          &        & KSDD2~\cite{KSDD}  & (93.5, 86.4, 94.2)  & (90.4, 82.9, 92.0)  & (81.7, 75.0, 85.5)  & (91.9, 84.5, 93.4)  & (94.9, 90.3, 96.2)  & \cellcolor[rgb]{ .906,  .902,  .902}(\best{98.0}, \best{93.9}, \best{98.3})  & \cellcolor[rgb]{ .906,  .902,  .902}(\secondbest{97.7}, \secondbest{93.1}, \secondbest{98.1}) \\
          &        & DAGM~\cite{DAGM}   & (91.8, 75.8, 79.5)  & (94.4, 80.3, 83.9)  & (62.1, 37.1, 32.3)  & (98.0, 90.6, 92.4)  & (\secondbest{98.3}, \secondbest{91.5}, \secondbest{94.2})  & \cellcolor[rgb]{ .906,  .902,  .902}(\best{99.5}, \best{95.0}, \best{97.8})  & \cellcolor[rgb]{ .906,  .902,  .902}(93.2, 83.9, 86.1) \\
          &        & DTD-Synthetic~\cite{DTD-Synthetic}  & (\secondbest{95.1}, 94.1, \secondbest{97.7})  & (85.5, 89.1, 94.0)  & (75.1, 86.3, 88.0)  & (93.7, 94.3, 97.4)  & (92.1, 92.4, 96.3)  & \cellcolor[rgb]{ .906,  .902,  .902}(\best{97.5}, \best{96.6}, \best{99.1})  & \cellcolor[rgb]{ .906,  .902,  .902}(91.0, \secondbest{94.4}, 97.4) \\
    \cmidrule{2-10}
          & \multicolumn{1}{c}{\multirow{6}[2]{*}{\makecell{Pixel-level\\(AUROC, $F_1$-max,\\ AP, PRO)}}}
          & MVTec-AD~\cite{MVTec}  & (82.3, 24.8, 18.2, 62.0)  & (87.5, 42.3, 39.1, 43.7)  & (77.9, 26.3, 21.1, 55.7)  & (\secondbest{91.0}, 38.9, 34.4, 81.7)  & (88.5, \secondbest{43.9}, 41.0, 47.6)  & \cellcolor[rgb]{ .906,  .902,  .902}(90.8, \secondbest{43.9}, \secondbest{41.2}, \secondbest{87.5})  & \cellcolor[rgb]{ .906,  .902,  .902}(\best{91.3}, \best{47.4}, \best{45.4}, \best{88.6}) \\
          &        & VisA~\cite{visa}   & (73.2, 9.0, 5.4, 51.1)  & (93.8, 32.6, 26.2, 86.5)  & (93.0, 24.1, 17.9, 80.2)  & (\secondbest{95.4}, 27.6, 20.7, 86.4)  & (95.1, 33.8, \secondbest{29.2}, 71.3)  & \cellcolor[rgb]{ .906,  .902,  .902}(\best{95.8}, \secondbest{34.6}, 28.4, \best{91.0})  & \cellcolor[rgb]{ .906,  .902,  .902}(95.3, \best{35.2}, \best{29.9}, \secondbest{88.2}) \\
          &        & BTAD~\cite{BTAD}   & (72.7, 18.5, 12.9, 27.3)  & (91.3, 40.1, 37.7, 21.0)  & (80.9, 24.1, 18.3, 41.4)  & (\secondbest{93.0}, \secondbest{47.1}, \secondbest{41.5}, 71.0)  & (87.7, 42.3, 36.6, 17.1)  & \cellcolor[rgb]{ .906,  .902,  .902}(91.1, \best{49.8}, \best{43.1}, \best{80.4})  & \cellcolor[rgb]{ .906,  .902,  .902}(\best{93.4}, 42.6, 38.7, \secondbest{76.7}) \\
          &        & KSDD2~\cite{KSDD}  & (94.1, 24.6, 17.4, 77.6)  & (94.5, \best{64.2}, \best{66.9}, 39.2)  & (95.6, 43.0, 39.6, 73.5)  & (98.0, 50.6, 43.7, 90.8)  & (96.1, 59.2, 58.6, 40.8)  & \cellcolor[rgb]{ .906,  .902,  .902}(\secondbest{98.5}, 60.9, 62.1, \secondbest{98.5})  & \cellcolor[rgb]{ .906,  .902,  .902}(\best{98.9}, \secondbest{64.0}, \secondbest{66.8}, \best{98.9}) \\
          &        & DAGM~\cite{DAGM}   & (87.6, 12.7, 6.8, 65.7)  & (84.4, 35.1, 27.8, 12.7)  & (69.1, 20.9, 14.7, 36.1)  & (\best{96.9}, \secondbest{56.9}, \secondbest{53.6}, \secondbest{89.2})  & (88.6, 48.5, 42.6, 37.6)  & \cellcolor[rgb]{ .906,  .902,  .902}(\secondbest{92.2}, \best{57.9}, \best{56.4}, \best{89.3})  & \cellcolor[rgb]{ .906,  .902,  .902}(89.5, 54.8, 52.2, 84.5) \\
          &        & DTD-Synthetic~\cite{DTD-Synthetic}  & (79.5, 16.1, 9.8, 51.5)  & (94.9, 60.4, 61.0, 33.8)  & (86.6, 35.8, 31.0, 63.2)  & (\secondbest{97.5}, 55.8, 52.5, \secondbest{87.9})  & (95.1, 58.4, 56.1, 34.3)  & \cellcolor[rgb]{ .906,  .902,  .902}(\best{98.1}, \secondbest{64.3}, \secondbest{65.5}, \best{94.8})  & \cellcolor[rgb]{ .906,  .902,  .902}(96.7, \best{65.8}, \best{67.7}, 92.4) \\
    \midrule
    \multirow{8}[4]{*}{Medical}
          & \multicolumn{1}{c}{\multirow{4}[2]{*}{\makecell{Image-level\\(AUROC, $F_1$-max,\\ AP)}}}
          & OCT17~\cite{OCT17}  & (55.2, 82.7, 81.1)  & (30.4, 81.7, 67.0)  & (58.1, 83.6, 84.1)  & (63.7, 85.7, 86.5)  & (77.3, 86.1, 91.8)  & \cellcolor[rgb]{ .906,  .902,  .902}(\secondbest{88.9}, \secondbest{88.2}, \secondbest{96.4})  & \cellcolor[rgb]{ .906,  .902,  .902}(\best{91.2}, \best{90.3}, \best{97.1}) \\
          &        & BrainMR1~\cite{BrainMRI}  & (86.6, 84.1, 91.5)  & (89.3, 88.2, 90.9)  & (83.0, 81.9, 89.3)  & (\secondbest{96.4}, \secondbest{94.7}, \secondbest{97.2})  & (94.9, 93.1, 96.8)  & \cellcolor[rgb]{ .906,  .902,  .902}(\best{96.7}, \best{94.8}, \best{97.6})  & \cellcolor[rgb]{ .906,  .902,  .902}(93.8, 90.3, 95.9) \\
          &        & Brain\_AD~\cite{Brain_AD01}  & (72.5, 90.7, 91.6)  & (58.8, 90.6, 87.8)  & (72.1, 91.3, 91.5)  & (69.0, 90.6, 90.1)  & (80.0, 90.2, 94.1)  & \cellcolor[rgb]{ .906,  .902,  .902}(\secondbest{80.8}, \secondbest{91.8}, \secondbest{94.7})  & \cellcolor[rgb]{ .906,  .902,  .902}(\best{87.1}, \best{92.5}, \best{96.7}) \\
          &        & HIS~\cite{HIS}  & (47.1, 45.6, 40.9)  & (59.8, 67.5, \best{56.5})  & (44.7, 66.6, 46.0)  & (55.2, 66.8, 56.1)  & (\secondbest{59.9}, 68.4, 55.8)  & \cellcolor[rgb]{ .906,  .902,  .902}(\best{60.1}, \best{68.9}, 55.9)  & \cellcolor[rgb]{ .906,  .902,  .902}(\best{60.1}, \secondbest{68.7}, \secondbest{56.2}) \\
    \cmidrule{2-10}
          & \multicolumn{1}{c}{\multirow{4}[2]{*}{\makecell{Pixel-level\\(AUROC, $F_1$-max,\\ AP, PRO)}}}
          & Brain\_AD~\cite{Brain_AD01}  & (87.6, 21.7, 13.3, 59.7)  & (83.6, 38.7, 35.0, 45.4)  & (94.1, 42.9, 40.5, 74.2)  & (95.1, 43.1, 42.3, 71.5)  & (\secondbest{95.2}, 40.6, 37.0, 36.4)  & \cellcolor[rgb]{ .906,  .902,  .902}(\secondbest{95.2}, \secondbest{46.7}, \secondbest{43.7}, \secondbest{79.5})  & \cellcolor[rgb]{ .906,  .902,  .902}(\best{96.4}, \best{50.2}, \best{51.9}, \best{83.5}) \\
          &        & CVC-ClinicDB~\cite{CVC-ClinicDB}  & (70.3, 27.3, 19.4, 32.5)  & (82.4, 38.2, 36.4, 53.5)  & (76.5, 32.1, 24.6, 52.8)  & (84.6, \secondbest{45.5}, \secondbest{41.7}, \secondbest{69.8})  & (84.3, 44.4, \best{43.3}, 63.6)  & \cellcolor[rgb]{ .906,  .902,  .902}(\secondbest{85.2}, \best{45.6}, 37.8, \best{70.6})  & \cellcolor[rgb]{ .906,  .902,  .902}(\best{85.9}, 45.0, 36.2, 67.3) \\
          &        & Endo~\cite{Endo}   & (68.2, 32.9, 23.8, 28.3)  & (82.7, 46.9, 47.2, 48.3)  & (78.1, 41.4, 36.8, 46.4)  & (\secondbest{86.5}, \secondbest{53.7}, \secondbest{50.7}, \secondbest{67.1})  & (82.9, 47.9, 47.7, 55.8)  & \cellcolor[rgb]{ .906,  .902,  .902}(84.9, 52.0, 46.2, 66.3)  & \cellcolor[rgb]{ .906,  .902,  .902}(\best{86.8}, \best{54.7}, \best{53.2}, \best{67.8}) \\
          &        & Kvasir~\cite{Kvasir}  & (69.7, 35.9, 27.8, 24.5)  & (77.6, 43.2, 42.5, 40.6)  & (73.6, 38.9, 31.1, 30.6)  & (\secondbest{82.0}, \secondbest{50.0}, 43.2, \secondbest{44.7})  & (80.3, 47.7, 31.0, 27.5)  & \cellcolor[rgb]{ .906,  .902,  .902}(80.3, 47.8, \best{47.6}, \best{50.8})  & \cellcolor[rgb]{ .906,  .902,  .902}(\best{82.6}, \best{50.3}, \secondbest{46.4}, 40.5) \\
    \bottomrule
    \end{tabular}%
  }
  \label{tab:result}%
\end{table*}%

\section{Experiments}

\subsection{Experimental Setup}
\textbf{Datasets.}  
To evaluate the zero-shot anomaly detection (ZSAD) performance of our model, we conduct experiments on 13 real-world datasets spanning both industrial and medical domains.  
In the industrial domain, we employ six widely used benchmarks: MVTec-AD~\cite{MVTec}, VisA~\cite{visa}, BTAD~\cite{BTAD}, KSDD2~\cite{KSDD}, DAGM~\cite{DAGM}, and DTD-Synthetic~\cite{DTD-Synthetic}.  
For the medical domain, we adopt OCT17~\cite{OCT17}, BrainMRI~\cite{BrainMRI}, BrainAD~\cite{Brain_AD01,Brain_AD02,Brain_AD03}, HIS~\cite{HIS}, CVC-ClinicDB~\cite{CVC-ClinicDB}, Endo~\cite{Endo}, and Kvasir~\cite{Kvasir}, covering diverse imaging modalities and tasks. Specifically, OCT17 contains retinal OCT scans for retinal disease classification, BrainMRI and BrainAD provide brain MRI images for brain tumor or lesion classification, and HIS consists of histopathology images for abnormal tissue analysis. CVC-ClinicDB, Endo, and Kvasir are colonoscopy datasets with pixel-level annotations for colon polyp segmentation.  Following previous works, we train the model on one industrial dataset and directly perform inference on other industrial and medical datasets without further fine-tuning.  
Since VisA contains object categories disjoint from the others, it is used as the auxiliary training set. For evaluating VisA itself, the model is fine-tuned on MVTec-AD. Detailed descriptions of all datasets are provided in Appendix.

\vspace{\baselineskip}
\noindent
\textbf{Evaluation metrics.}  
For classification, we report image-level metrics including the Area Under the ROC Curve (AUROC), the maximum $F_1$ score over all thresholds ($F_1$-max), and Average Precision (AP).
For segmentation, we evaluate pixel-level AUROC,  the maximum $F_1$ score over all thresholds ($F_1$-max), AP, and Per-Region Overlap (PRO) to assess the localization capability of the model.

\vspace{\baselineskip}
\noindent
\textbf{Implementation Details.} We adopt publicly available backbones: CLIP (ViT-L/14@336px) and DINOv2 (ViT-L/14). Following standard practice, input images are resized to $518\times518$. From the 24-layer visual encoder, we extract patch tokens at layers $\{6,12,18,24\}$. The number of anchor queries in SCA is set to $m=4$ by default. VisualAD is trained with the Adam optimizer, using an initial learning rate of $1\times10^{-3}$ and a batch size of $8$. Additional implementation details are provided in Appendix.
\begin{figure}[!htb]
    \centering
    \includegraphics[width=0.5\textwidth]{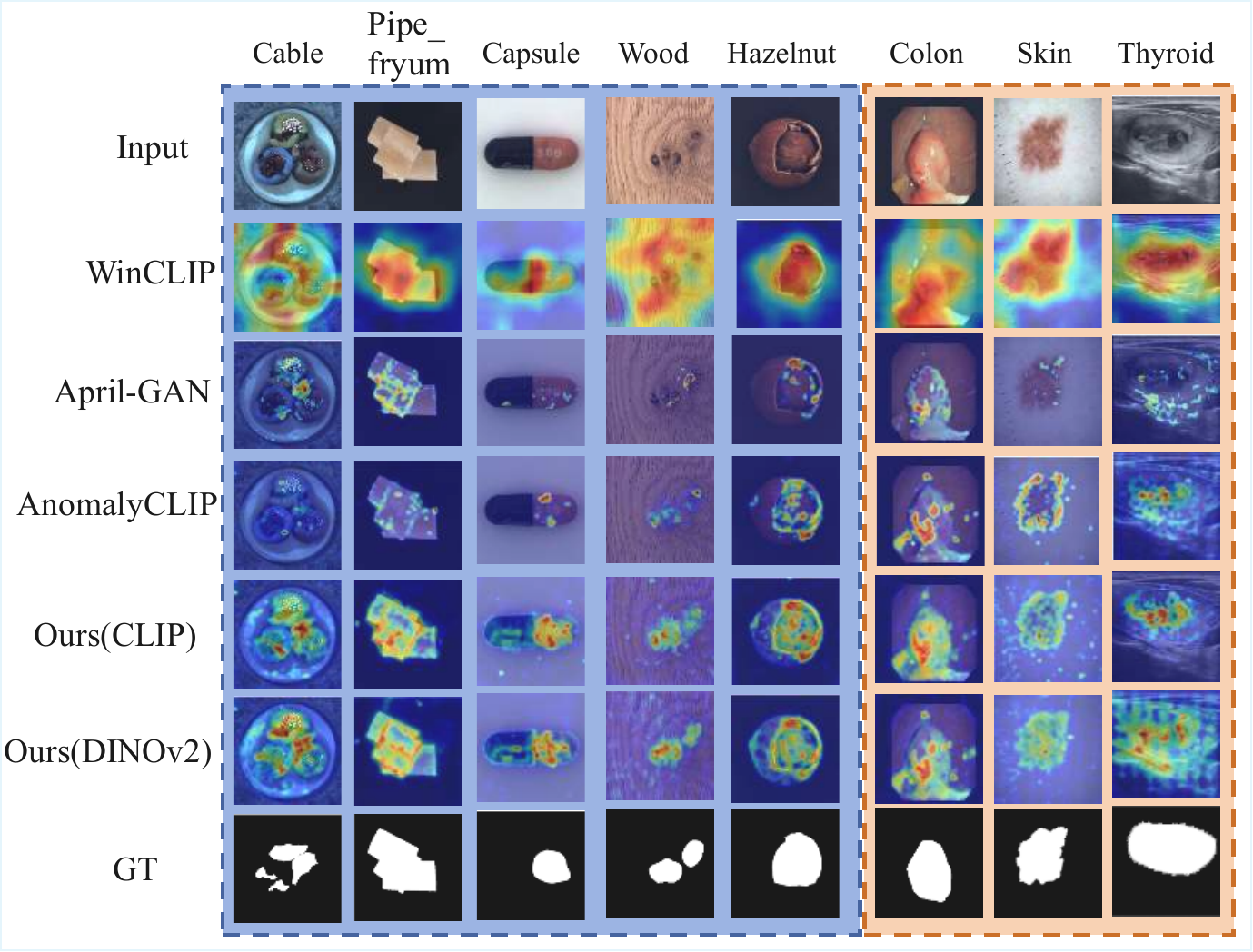} 
    \caption{Qualitative comparison of anomaly segmentation results for different ZSAD methods. The first five columns show images from industrial datasets, while the last three columns correspond to medical datasets.}
    \label{fig:result}
\end{figure}

\subsection{Performance Comparison with SOTA Method}
In this study, we compare VisualAD with five state-of-the-art methods: WinCLIP~\cite{WinCLIP}, APRILGAN~\cite{APRIL-GAN}, CLIP-AD~\cite{CLIP-AD}, AnomalyCLIP~\cite{AnomalyCLIP}, and AdaCLIP~\cite{AdaCLIP}. Since our approach can adapt to multiple backbones, we report results using both CLIP’s ViT-L/14@336px and DINOv2 (ViT-L/14) to demonstrate its flexibility across different visual architectures. Table \ref{tab:result} reports quantitative ZSAD results on industrial and medical benchmarks. Compared with existing methods, \textbf{VisualAD} achieves state-of-the-art performance on almost all datasets at both the image and pixel levels. On industrial datasets in particular, VisualAD with CLIP ViT-L/14@336px attains the best results across all classification metrics. We further observe that CLIP-based and DINOv2-based instances of VisualAD exhibit complementary strengths, with the former slightly favoring image-level classification and the latter delivering stronger pixel-wise segmentation performance. Fig.~\ref{fig:result} visualizes anomaly maps from representative industrial and medical datasets. Compared with other approaches, VisualAD produces more accurate segmentations and more complete localization of anomalous regions. The advantage is especially pronounced on medical categories (for example, skin), where our method yields clearer boundaries and fewer false positives. These results indicate that high-level anomaly semantics learned purely in the visual feature space can generalize effectively across domains.

\subsection{Ablation Studies}
In this section, we use CLIP’s ViT-L/14@336px backbone by default and conduct ablation studies on the VisA dataset to examine how different settings affect the proposed VisualAD method. We additionally include DINOv2-based ablations for part of the experiments, with the full results deferred to the appendix.
\paragraph{Effect of Different Components.}
\begin{table}[htbp]
  \centering
  \caption{Ablation study on different modules and loss components using ViT-L/14@336px. Both Image-level and Pixel-level metrics are reported as (AUROC, AP).}
  \label{table:ablation}
  \setlength{\tabcolsep}{8pt}
  \resizebox{0.5\textwidth}{!}{%
  \begin{tabular}{c c c c c c c c}
    \toprule
    \textbf{Ablation} & \textbf{SCA} & \textbf{SAF} & \textbf{Focal} & \textbf{Dice} & \textbf{Ctr} & \textbf{Image-level} & \textbf{Pixel-level} \\
    \midrule
    \multirow{3}{*}{\textbf{Module}} 
      & \xmark & \cmark & \cmark & \cmark & \cmark & (82.3, 85.4) & (95.3, 27.4) \\
      & \cmark & \xmark & \cmark & \cmark & \cmark & (50.5, 58.9) & (87.9, 3.5) \\
      & \xmark & \xmark & \cmark & \cmark & \cmark & (48.0, 56.1) & (38.3, 0.8) \\
    \midrule
    \multirow{3}{*}{\textbf{Loss}} 
      & \cmark & \cmark & \xmark & \cmark & \cmark & (83.5, 86.8) & (95.3, 26.6) \\
      & \cmark & \cmark & \cmark & \xmark & \cmark & (83.4, 86.5) & (95.3, 27.3) \\
      & \cmark & \cmark & \cmark & \cmark & \xmark & (84.2, 87.0) & (95.4, 27.6) \\
    \midrule
    \rowcolor{gray!12}
    \textbf{VisualAD} 
      & \textbf{\cmark} & \textbf{\cmark} & \textbf{\cmark} & \textbf{\cmark} & \textbf{\cmark} 
      & \textbf{(84.7, 87.6)} & \textbf{(95.8, 28.4)} \\
    \bottomrule
  \end{tabular}
  }
  \label{tab:ablation}
\end{table}
\begin{figure}[!htb]
    \centering
    \includegraphics[width=0.5\textwidth]{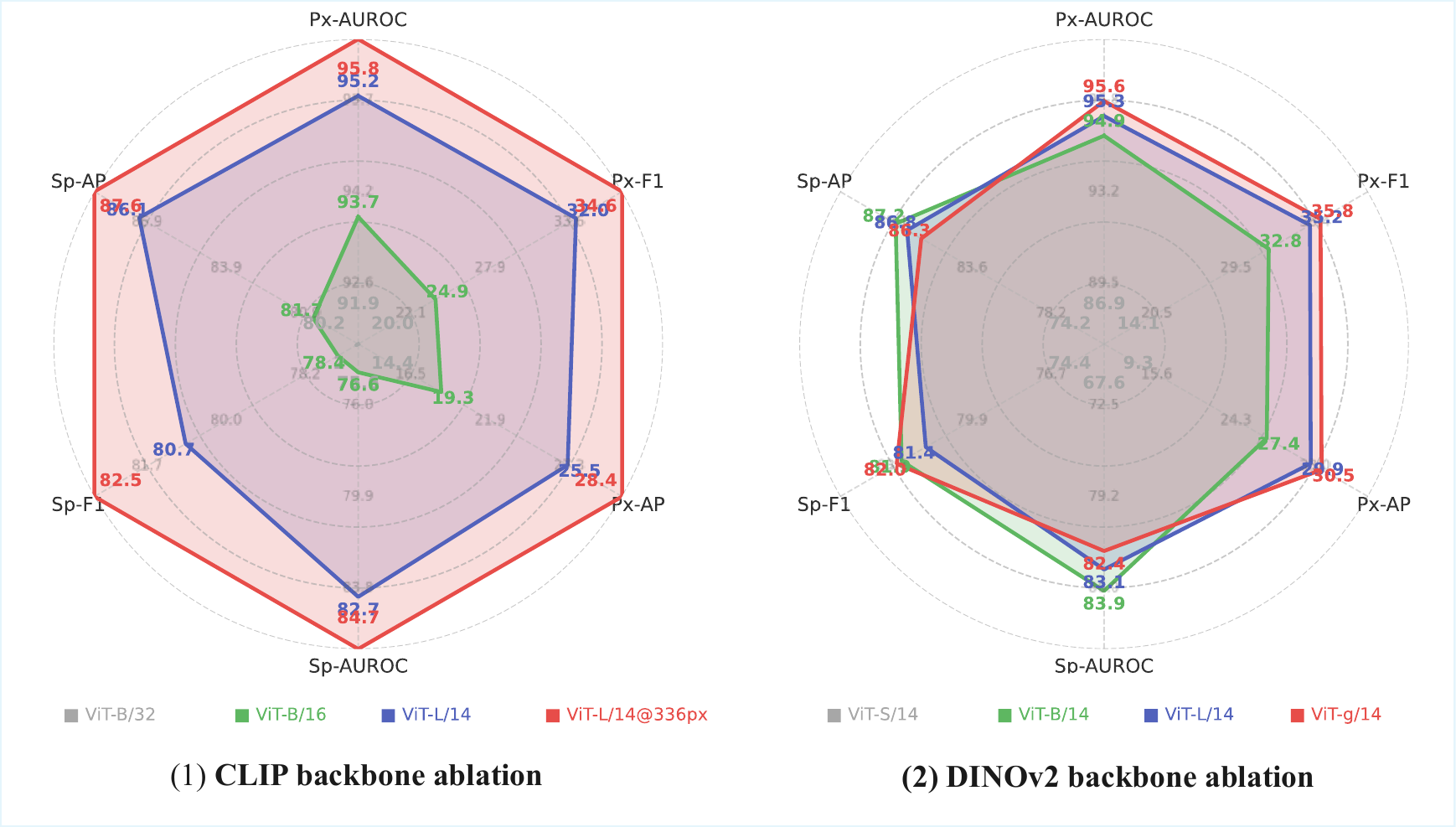} 
    \caption{Radar plot comparing different pretrained ViT backbones under the ZSAD setting.
We evaluate CLIP and DINO-based variants from small to large model scales.
Larger backbones and higher input resolutions generally lead to consistent improvements in both pixel-level and sample-level metrics,
with CLIP’s ViT-L/14@336px giving the strongest sample-level accuracy and DINO’s ViT-g/14 delivering the highest pixel-level precision.}
    \label{fig:back}
\end{figure}
\begin{table*}[htbp]
\centering

\begin{minipage}{0.48\textwidth}
\centering
\caption{Ablation on the number of anchor queries using a ViT-L/14@336px backbone pre-trained with CLIP.
Pixel-level and image-level metrics are reported as (AUROC, $F_1$-max, AP).
The best and second-best results are highlighted in \textbf{bold} and \underline{underline}, respectively.
}
\vspace{2mm}
\begin{tabular}{ccc}
\toprule
\textbf{Anchors} & \textbf{Pixel-level} & \textbf{Image-level} \\
\midrule
1  & (\underline{95.2}, 33.3, 27.0) & (\underline{84.7}, 82.2, 87.3) \\
2  & (\underline{95.2}, 33.4, \underline{27.2}) & (83.8, 81.3, 86.7) \\
\rowcolor[rgb]{ .906, .902, .902}
4  & (\textbf{95.8}, \textbf{34.6}, \textbf{28.4}) 
   & (\underline{84.7}, \underline{82.5}, \underline{87.6}) \\
8  & (95.0, 32.9, 26.8) & (84.0, 81.7, 86.7) \\
16 & (95.1, \underline{33.5}, 26.9) & (\textbf{84.9}, \textbf{83.0}, \textbf{87.8}) \\
32 & (94.6, 32.4, 26.1) & (84.5, 82.1, \underline{87.6}) \\
\bottomrule
\end{tabular}
\label{tab:anchor_num}
\end{minipage}
\hfill
\begin{minipage}{0.48\textwidth}
\centering
\caption{Ablation on different layer combinations using ViT-L/14@336px backbone pre-trained with CLIP.
Pixel-level and image-level metrics follow (AUROC, $F_1$-max, AP).Best and second-best results are in \textbf{bold} and \underline{underline}.}
\vspace{2mm}
\begin{tabular}{ccc}
\toprule
\textbf{Layers} & \textbf{Pixel-level} & \textbf{Image-level} \\
\midrule
\{6\}               & (88.0, 11.9, 8.3)   & (67.1, 75.0, 72.7) \\
\{12\}              & (94.7, 28.4, 22.9)  & (76.6, 77.3, 81.5) \\
\{18\}              & (\underline{95.2}, \underline{31.3}, \underline{24.6})  
                    & (\underline{82.4}, \underline{80.5}, \underline{85.4}) \\
\{24\}              & (93.5, 25.9, 19.3)  & (79.6, 79.1, 84.0) \\
\{6, 12\}           & (93.7, 29.3, 24.1)  & (79.5, 78.5, 83.6) \\
\{6, 12, 18\}       & (95.0, 32.9, 27.1)  & (83.4, 80.5, 86.4) \\
\rowcolor{gray!15}
\{6, 12, 18, 24\}   & (\textbf{95.8}, \textbf{34.6}, \textbf{28.4}) 
                    & (\textbf{84.7}, \textbf{82.5}, \textbf{87.6}) \\
\bottomrule
\end{tabular}
\label{tab:layer_ablation}
\end{minipage}

\end{table*}

We conduct separate ablations on modules and losses, with results summarized in Table~\ref{tab:ablation}. 
For the module ablation, removing any single component consistently degrades performance, indicating that each module contributes positively to the framework. 
Eliminating SCA weakens spatial alignment and fine-grained evidence aggregation, while removing the SAF undermines nonlinear recalibration of high-level semantics; dropping both leads to the most pronounced degradation, highlighting their complementary roles. 
For the loss ablation, excluding the Focal, Dice, or Ctr objective each yields a noticeable decline, suggesting that the hybrid supervision balances classification focus, region consistency, and feature separability. 
Overall, the full model achieves the best image-level and pixel-level results, validating the effectiveness and synergy of all components.

    \paragraph{Ablation on different pretrained ViT.} We evaluate both CLIP and DINO pretrained ViTs from small to large capacities and observe a consistent trend: scaling the backbone and input resolution steadily improves pixel- and image-level performance, with gains tapering at the largest scales. Within CLIP, moving from ViT-B to ViT-L with 336px inputs yields the strongest overall results, while within DINO the larger variants approach or surpass smaller CLIP models in pixel-level {$F_1$-max} and AP. Notably, within the CLIP family ViT-L/14@336px attains the best sample-level results, whereas within the DINO family ViT-g/14 achieves the best pixel-level precision (highest pixel {$F_1$-max}). The Fig.~\ref{fig:back} mirrors these patterns with broadly rising curves across AUROC, {$F_1$-max}, and AP.

\paragraph{Influence of the number of anchors.}
Table~\ref{tab:anchor_num} analyzes how the number of anchor queries in SCA affects model performance. At the pixel level, increasing the anchors from 1 to 4 yields steady improvements across AUROC, $F_1$-max, and AP, with \(m=4\) delivering the strongest overall results. Expanding the anchor set beyond 4 (e.g., 8--32) introduces redundancy and slightly degrades pixel-level accuracy, indicating that an excessively large anchor set dilutes spatial focus rather than enhancing it. Image-level metrics exhibit a flatter trend: while \(m=16\) provides the highest AUROC, $F_1$-max, and AP, the improvements over \(m=4\) remain within one to two percentage points. These findings suggest that a small anchor set is sufficient, and we adopt \(m=4\) as the default choice, achieving a balanced trade-off between pixel-level precision and image-level robustness.

\paragraph{Influence of ensemble of different patch-level image layers.}
Table~\ref{tab:layer_ablation} examines the effect of aggregating features from different intermediate layers of the frozen image encoder. Using a single layer (\{6\}, \{12\}, \{18\}, or \{24\}) shows that mid-level layers yield stronger representations than shallow or deep ones, with layer \{18\} achieving the best balance between local detail and global semantics. When combining multiple layers, the performance improves consistently: \{6,12,18\} already brings clear gains in both pixel- and sample-level metrics, and extending to \{6,12,18,24\} further enhances stability and overall accuracy. These results demonstrate that multi-layer feature ensembles effectively capture complementary information across scales, leading to more robust anomaly localization and classification.
\begin{table}[htbp]
  \centering
  \caption{Effect of positional modeling in the SCA module on VisA.
  Metrics are reported as (AUROC, $F_1$-max, AP). Best and second-best
  results are highlighted in \textbf{bold} and \underline{underline}.}
  \footnotesize
  \setlength{\tabcolsep}{6pt}
  \renewcommand{\arraystretch}{1.5}
  \begin{tabular}{ccc}
    \toprule
    \textbf{Positional modeling} & \textbf{Pixel-level} & \textbf{Image-level} \\
    \midrule
    None (no PE)               
        & (93.0, 29.0, 22.5) 
        & (82.4, 78.1, 84.5) \\
    Learnable 1D Absolute      
        & (\textbf{95.8}, \textbf{34.6}, \textbf{28.4})
        & (\textbf{84.7}, \textbf{82.5}, \textbf{87.6}) \\
    Fixed 1D Sinusoidal        
        & (\underline{95.2}, \underline{33.8}, \underline{27.6})
        & (\underline{84.1}, \underline{81.9}, \underline{86.8}) \\
    Learnable 2D Absolute      
        & (95.0, 33.2, 27.0)
        & (83.8, 81.6, 86.0) \\
    Fixed 2D Sinusoidal        
        & (94.6, 32.4, 26.1)
        & (83.4, 80.8, 85.4) \\
    Learnable 2D Relative Bias 
        & (94.4, 31.9, 25.7)
        & (83.0, 80.2, 84.9) \\
    \bottomrule
  \end{tabular}
  \label{POS}
\end{table}

\begin{figure*}[!htb]
    \centering
    \includegraphics[width=1\textwidth]{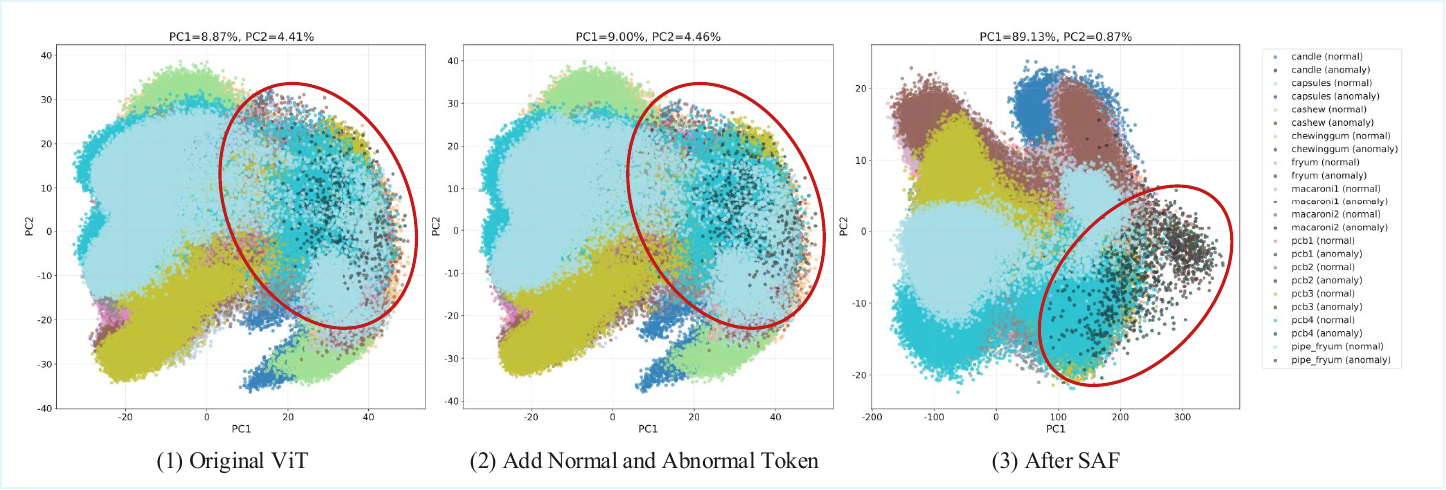} 
    \caption{PCA visualization of feature distributions under three configurations.
From left to right: vanilla CLIP features, CLIP with normal/anomaly tokens, and CLIP with both tokens and the MLP-based transformation.
The anomaly cluster (dark) becomes progressively more compact and moves farther from the normal cluster, while the variance concentrates along a single dominant axis, indicating increasingly discriminative representations for anomaly separation.}
    \label{fig:feature}
\end{figure*}
\subsection{Visualization}
\begin{figure}[!htb]
    \centering
    \includegraphics[width=0.5\textwidth]{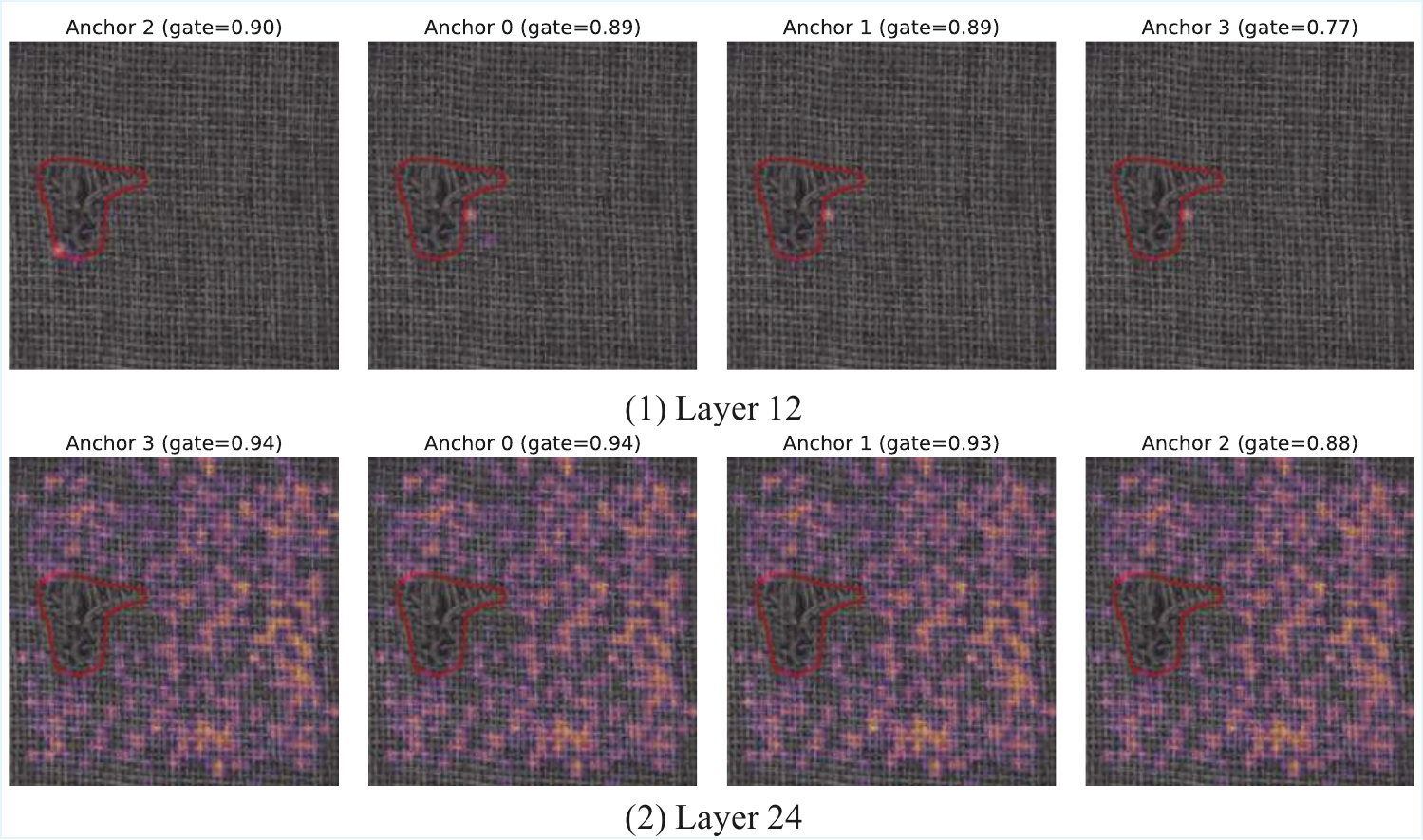} 
    \caption{Layer-wise anchor attention maps on the \textit{carpet} category.
Anchors within the same layer attend to similar regions, whereas attention patterns differ across layers: mid-level layers focus on fine-grained defect boundaries, while deeper layers emphasize broader normal regions.
This complementarity between layers provides both local anomaly cues and holistic normal context for accurate localization.}
    \label{fig:anchor}
\end{figure}
\paragraph{Influence of spatial positional modeling.}
Although we re-train the positional encoding associated with the inserted global tokens, we still observe that their interaction with patch tokens tends to blur the underlying spatial geometry and degrade pixel-level localization. To improve spatial consistency, we equip the SCA module with explicit positional modeling and conduct an ablation study comparing six variants: no positional encoding, learnable 1D absolute embeddings, fixed 1D sinusoidal encodings, learnable 2D absolute embeddings, fixed 2D sinusoidal encodings, and learnable 2D relative positional bias. As summarized in Table~\ref{POS}, we report both pixel-level and image-level performance in the format (AUROC, $F_1$-max, AP). A simple learnable 1D absolute embedding achieves the best overall trade-off across these metrics, while more complex 2D or relative formulations offer only marginal gains or even harm stability. Therefore, we adopt the learnable 1D absolute variant as the default configuration in the SCA module.

We provide visual analyses to illustrate how anchor attention and feature representations evolve within VisualAD.
Fig.~\ref{fig:anchor} shows layer-wise anchor attention on the \textit{carpet} category.
Within each layer, anchors consistently attend to nearly identical spatial regions, revealing stable and coherent local evidence.
Across layers, the focus shifts in a structured manner: mid-level layers (for example, layer~12) emphasize fine-grained defect boundaries and subtle texture disruptions, whereas deeper layers (for example, layer~24) highlight broader, predominantly normal regions.
This progression reflects the hierarchical nature of the frozen ViT encoder, where mid-level anchors capture detailed irregularities while high-level anchors encode holistic normal patterns, producing complementary cues that strengthen anomaly localization.
Together with the quantitative ablations, these attention maps clearly demonstrate that the anchor-based SCA module leverages multi-layer information rather than relying on a single feature scale.

To examine representation evolution, Fig.~\ref{fig:feature} plots PCA distributions under three configurations.
Vanilla CLIP features exhibit substantial overlap between normal and anomalous samples, with PC1 accounting for only 8.8\%, suggesting poor separability.
Introducing learnable normal and anomaly tokens slightly increases PC1 to 9.0\%, indicating a weak emerging discriminative direction.
After applying SAF, PC1 rises to 89.1\%, with variance concentrated along a dominant axis that cleanly separates the two classes.
The anomaly cluster also becomes more compact and shifts farther from the normal cluster, as highlighted by the red circles.
These trends show that the tokens provide the initial discriminative structure, while SAF strengthens intra-class compactness and enlarges the inter-class margin.

\section{Conclusion}
\label{conclusion}
VisualAD introduces a purely visual approach to zero-shot anomaly detection by inserting two learnable global tokens (normal and abnormal) into a frozen Vision Transformer and grounding them with spatial evidence via Spatial Cross-Attention and a Self-Alignment Function. Trained only on auxiliary industrial data and without any text branch or cross-modal alignment, the method generalizes directly to unseen industrial and medical categories, delivering state-of-the-art image- and pixel-level performance. Ablations confirm that token learning, the SCA/SAF design, and multi-layer fusion each contribute indispensably to detection and localization. Future directions include adaptive layer selection, richer structured tokens for finer granularity, and lightweight uncertainty estimation to improve robustness under strong domain shift.
\section{Acknowledgments} 
This work was supported in part by the National Natural Science Foundation of China (No. 62576001, No. 62206003).

{
    \small
    \bibliographystyle{ieeenat_fullname}
    \bibliography{mainn}
}

\end{document}